\def\year{2019}\relax
\documentclass[letterpaper]{article} 
\usepackage{aaai19}  
\usepackage{times}  
\usepackage{helvet}  
\usepackage{courier}  
\usepackage{url}  
\usepackage{graphicx}  

\newtheorem{definition}{Definition}

\usepackage{booktabs} 
\usepackage[noend]{algpseudocode}
\usepackage[ruled,vlined]{algorithm2e}
\usepackage{subfigure}
\usepackage{color}
\usepackage{amsmath}
\usepackage{caption}
\usepackage[shortlabels]{enumitem}

\begin{document}
%
\title{Predicting Urban Dispersal Events: A Two-Stage Framework through Deep Survival Analysis on Mobility Data}
\author{Amin Vahedian \ Xun Zhou \ Ling Tong \ W. Nick Street\\ 
The University of Iowa\\
\{amin-vahediankhezerlou,xun-zhou,ling-tong,nick-street\}@uiowa.edu\\
\ And\\
\textbf{Yanhua Li}\\
Worcester Polytechnic Institute\\
yli15@wpi.edu
}
\maketitle
\begin{abstract}
Urban dispersal events are processes where an unusually large number of people leave the same area in a short period. Early prediction of dispersal events is important in mitigating congestion and safety risks and making better dispatching decisions for taxi and ride-sharing fleets. 
Existing work mostly focuses on predicting taxi demand in the near future by learning patterns from historical data. However, they fail in case of abnormality because dispersal events with abnormally high demand are non-repetitive and violate common assumptions such as smoothness in demand change over time.
Instead, in this paper we argue that dispersal events follow a complex pattern of trips and other related features in the past, which can be used to predict such events. Therefore, we formulate the dispersal event prediction problem as a survival analysis problem. We propose a two-stage framework (DILSA), where a deep learning model combined with survival analysis is developed to predict the probability of a dispersal event and its demand volume. We conduct extensive case studies and experiments on the NYC Yellow taxi dataset from 2014-2016. Results show that DILSA can predict events in the next 5 hours with F1-score of $0.7$ and with average time error of $18$ minutes. It is orders of magnitude better than the state-of-the-art deep learning approaches for taxi demand prediction. 
\end{abstract}

\section{Introduction}
An urban dispersal event is the process where an abnormally large crowd leaves the same area within a short period. Dispersal events can be observed after large gathering events, such as concerts, sporting events, or protests. Unexpected dispersal events often cause public safety risks, congestion, and high demands of public transportation resources (e.g., taxis) within a short period. Therefore, early prediction of large dispersal events as well as the crowd size are of great importance to a number of different parties. (1) Public safety officials and traffic administrators can benefit from such a technique since they could allocate resources and make plans to mitigate potential risks or congestion. (2) Transportation stakeholders such as taxi drivers and fleet managers are enabled to improve profit by dispatching more taxis to such events if they can be predicted in advance.    

Thus, dispersal event prediction is {\bf non-trivial} and {\bf necessary}. While most of the large events have schedules, the time of dispersion often has a high level of uncertainty due to varying occasions, attendees, and other factors such as weather. Moreover, many events are not planned or have much higher attendance than expected, such as social protests and gatherings. In addition, many events are not public and only known to special interest groups and it is not possible for the public to collect schedules of such events. For example, large groups of Pokemon Go game players gather for special events in the game. Social activities organized through instant messaging tools are not public, either. Finally, collecting and verifying schedule information from various sources often requires costly human labor and cannot be done in a fully automated manner.

In recent years many big mobility datasets such as taxi trip data and For-Hire Vehicle (FHV) requests (e.g., Uber), have become available. Automatic dispersal event prediction, therefore, has become feasible. A typical solution is to build models to predict taxi demand using the above datasets and identify high-demand locations as dispersal events. Related research shows taxi demand has a highly predictable pattern, when predicting near future \cite{zhao2016predicting,xu2017real,zhang2016framework,zhao2016predicting,moreira2013predicting,davis2016multi}. However, that is often true only for regular demand prediction rather than abnormally high demand. In such a case, the assumptions of pattern repetitiveness are violated and the methods will fail to provide a timely and accurate prediction. In particular, their ability to make long-term forecast of abnormally high demand is weak due to assumptions that demand changes smoothly (auto-correlated) over time. 

In this paper, we focus on predicting such non-periodic and unexpected large dispersal events with abnormally high taxi demand. Specifically, given the historical taxi trip records and other relevant features (e.g., weather, POI), we predict (1) when and where abnormally high taxi demand will occur, and (2) the volume of demand in the predicted time span of the dispersal event. 

\begin{figure}
	\centering
	\subfigure[\small An anomalously high pick-up, preceded by an anomalously high number of arrivals.]{\includegraphics[width=0.2\textwidth]{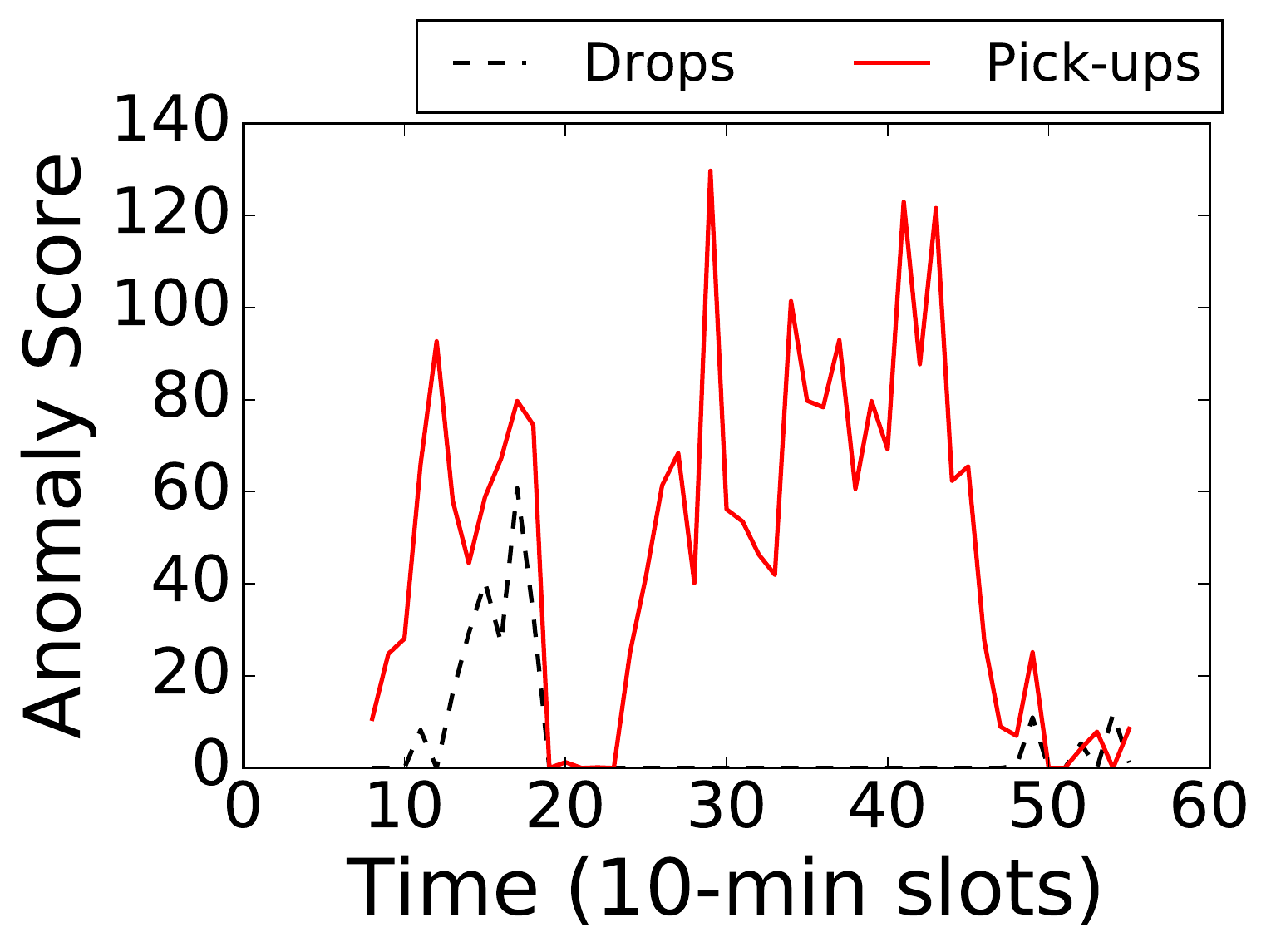}}
	\hfil
	\subfigure[\small An anomalously high pick-up, not preceded by a drop event.]{\includegraphics[width=0.2\textwidth]{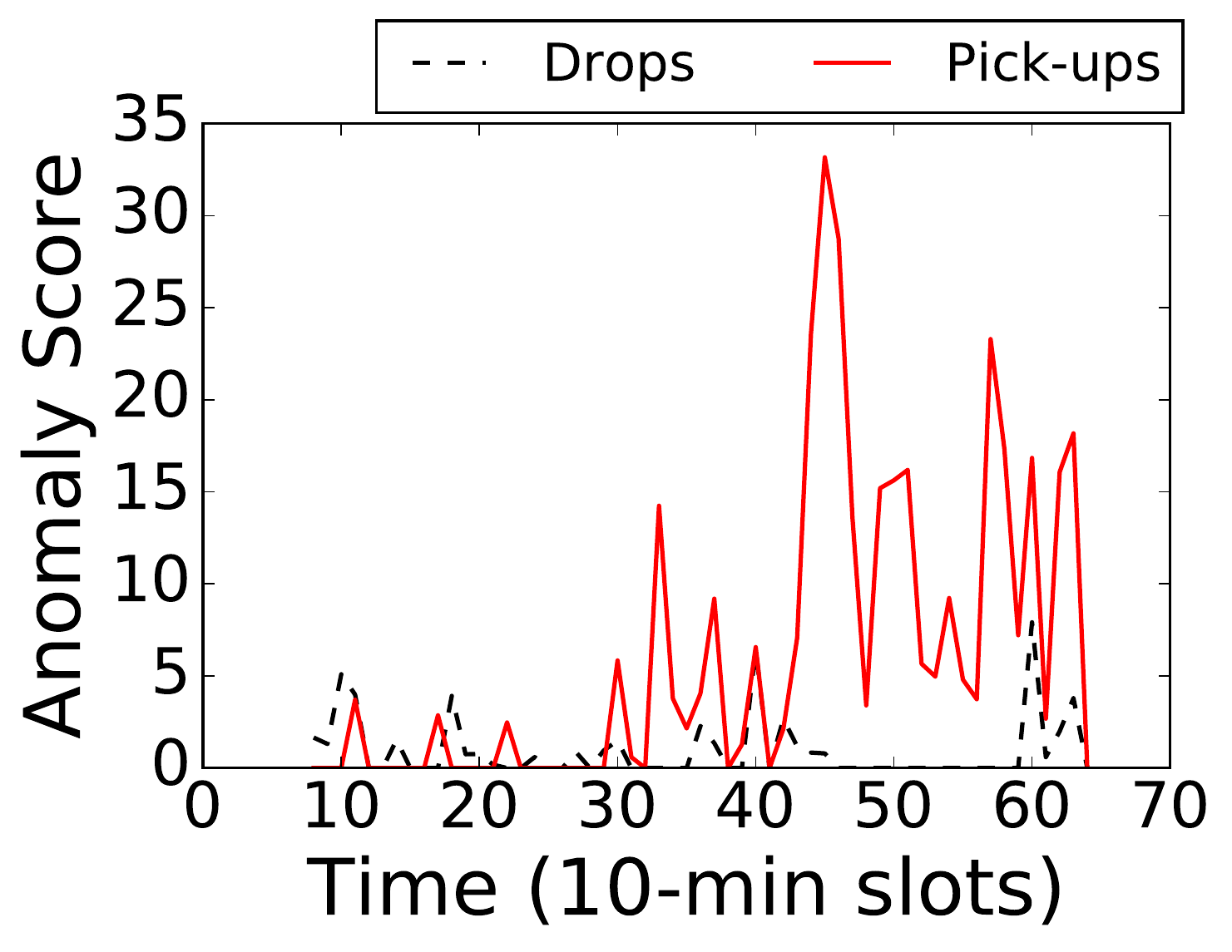}}
	\caption{\small Examples of abnormally large number of pick-ups.}
	\label{example_1}
\end{figure}

To address the limitations of related work, we propose an alternative solution. Firstly, we treat dispersal event prediction as a ``Survival Analysis'' problem, where we learn a model to predict the probability of ``death'' (event occurs) at each location in the future. Secondly, we argue that there is evidence of demand abnormality during the time leading to it, which can be used to train the survival analysis model. The intuition is that dispersal events are often caused by some forms of gatherings, which can indicate future abnormally high demand. Figure \ref{example_1} (a) shows an example of abnormally high pick-ups for a concert event at Madison Square Garden in New York City. The dashed and solid lines represent the anomaly scores  \cite{neill2009expectation} of the drops and the pick-ups, respectively. The large anomaly in pick-ups towards the end of the event follows an anomaly in the drops earlier. 
However, such patterns are not always as explicit. Fig. \ref{example_1} (b) shows such a case around McKittrick Hotel in Manhattan, where there are no abnormal drops preceding the dispersal event. In such cases, finding the right signals for predicting dispersal events is more challenging.

In this paper, we make the following novel contributions: \textbf{(1) We propose a two-stage framework}, using deep neural networks to predict dispersal events. We incorporate various features including spatial, temporal, weather, and Point of Interest (POI) features, in addition to recent taxi pick-up and drop observations. 
\textbf{(2) We formulate the dispersal event prediction problem as a survival analysis problem} \cite{miller2011survival}.
In the two-stage prediction framework, we predict the time of abnormal demand using survival analysis and then predict the demand volume. We call our method \textbf{DILSA}, DIspersaL event prediction using Survival Analysis. We evaluate our methods using real-world data from New York City. Our evaluations show our method identifies dispersal events with F1-score of $0.7$, while the error for predicting the time of the dispersal event is $18$ minutes for a $5$-hour prediction period. Also, our method predicts the pick-up demand in case of anomaly with superior accuracy compared to a baseline.

The rest of the paper is organized as follows: In the next section we discuss the related work, followed by problem formulation and our proposed computational solution. Then, we present the evaluations and conclude the paper.

\section{Related Work}
\label{related_work}
Prior related work include (1) event detection and forecasting, (2) taxi demand prediction, and (3) survival analysis. 

\textbf{Event Detection and Forecasting:} Event detection has been widely studied in various domains, including public health, urban computing, and social network analysis. The works \cite{kulldorff1997spatial,kulldorff2005space,neill2009expectation} and other recent works on event detection \cite{li2012mining,hong2015detecting} use already observed counts. An event is defined as a region with significantly higher counts, such as disease reports or number of taxi drops. Social media posts and geo-tagged tweets have been used as well to detect and forecast events such as social unrests and protests~\cite{zhou2014event,chen2014non,liu2016graph,zhang2017triovecevent}. Regions and time windows where the frequency of certain keywords exhibit abnormal changes are identified as events. These works do not use mobility data. The \textbf{dynamic patterns} of the events such as gathering or dispersing \textbf{are not captured}.

Works \cite{zhou2016traffic,khezerlou2017traffic,hoang2016fccf} use traffic flow data to detect gathering events. Vahedian et al. use destination prediction to predict gathering events \cite{Vahedian2017forecasting}. However, \textbf{such methods are not applicable} to dispersal events, as trajectories and traffic flow are \textbf{observed only after such events}.

\textbf{Taxi demand prediction} has been studied closely in recent years, due to access to public taxi datasets \cite{zhao2016predicting}. To the best of our knowledge none of the proposed methods directly address the prediction problem in case of anomaly. State-of-the-art methods for predicting taxi demand use historical data and time series analysis. \cite{yao2018deep} propose a deep learning framework which captures the spatial and temporal dependencies to predict taxi demand. \cite{xu2017real} formulate an LSTM Network to learn the regular pattern of taxi demand. \cite{zhao2016predicting} show the regular taxi demand is highly predictable and test different algorithms to approach the maximum accuracy. \cite{zhang2016framework} used spatial clustering to predict demand hotspots. They predict areas with high density of demand using DBSCAN. Such areas, despite having high demand, are part of the regular pattern. \cite{moreira2013predicting} used streams of taxi data as time series to predict taxi demand in the next 30-minute period. \cite{davis2016multi} used time series analysis to solve the demand prediction problem, giving recommendations to drivers. \cite{mukai2012taxi} used a simple multi-output ANN to predict demand, using features created from recent demand, time and weather information.

The above-mentioned research aims at learning the \textbf{regular pattern} of taxi demand in absence of anomaly. Considering the regular demand is highly predictable, in this paper, we take on the harder \textbf{challenge of predicting anomalous taxi demand}, which we believe is of greater importance.

\textbf{Survival analysis} is the analysis of duration of time until an event. It has been applied in engineering as well as health practices \cite{street1998neural}, for which it was originally developed \cite{miller2011survival}. To the best of our knowledge, this is the first time survival analysis is used in the context of urban event prediction. In this paper, we propose to use a deep Artificial Neural Network to predict the probabilities of survival. Predicting the probabilities of survival at different time points using a common internal representation (the hidden nodes of a deep ANN) allows the learned model to share information across the time points, resulting in better predictive results.

\section{Problem Formulation}
\label{sec_formulation}
\subsection{Concepts and Definitions}
\label{sec_def}
We define a spatio-temporal field $Z=(S,T)$ as a two-dimensional geographical region $S$ paired with a period of time $T$. $S$ is partitioned by a grid. Each grid cell $l_1,l_2,...,l_{|S|}$ represents a distinct location in the geographical region. $T$ is partitioned into fixed-length time-steps. Given $Z$, the location of any moving object, can be mapped into a grid cell in $S$ and a time-step in $T$. For instance, pick-up and drop location and time of a taxi trip can be represented by $(l^s,t^s,l^d,t^d)$, where $(l^s,t^s)$ are source location and time and $(l^d,t^d)$ correspond to the destination. \textbf{Pick-up count} $\big(C^p_{l,t}\big)$ of grid cell $l$ at time $t$ is the number of trips with source $(l,t)$. Similarly, \textbf{drop count} $\big(C^d_{l,t}\big)$, is the number of trips with destination $(l,t)$. Since the drop and pick-up counts demonstrate a periodic pattern, we define baseline counts to represent the expected counts. \textbf{Pick-up baseline} $\big(B^p_{l,t}\big)$ of grid cell $l$ at time $t$ is the average of pick-up counts at $l$ at the same time of day. \textbf{Drop baseline} $\big(B^d_{l,t}\big)$ is defined similarly. A \textbf{spatio-temporal region} $R=(S_R,T_R)$ is a rectangular sub-field of $S$ paired with a continuous subset of $T$. The counts and baselines can be obtained for any spatio-temporal region, defined bellow. To study the abnormally high taxi demands, in this paper, we are interested in regions, where there are significantly higher counts than expected, i.e., when $C^p_R$ is significantly higher than $B^p_R$. We assume $C^p_R$ follows a Poisson distribution and test the following hypotheses: $H_0$: $C^p_R$ is from a Poisson distribution of parameter $B^p_R$, $H_1$: $C^p_R$ is from a Poisson distribution of a parameter larger than $B^p_R$. We use the Expectation-based Likelihood Ratio Test of \cite{neill2009expectation}:
\begin{equation}
\small
\label{ebp}
LLR(R) = \begin{cases}
C^p_R\log\frac{C^p_R}{B^p_R}+(B^p_R-C^p_R) & \text{if\ } C^p_R\geq B^p_R \\
0 & otherwise
\end{cases}
\end{equation}
\cite{zhou2016traffic} showed that $LLR(R)$ is at $\alpha$-level significance if $1-Pr(X\leq C^p_R)\leq \alpha$, where $X\sim Poisson(B^p_R)$. Therefore, we define dispersal events:

\begin{definition}
	There is a \textbf{dispersal event} at spatio-temporal region $R$, if $LLR(R)$ is significant at $\alpha$-level.
	\label{disE}
\end{definition}

Locations have specific attributes other than the pick-up and drop counts. We consider two of them: weather and Point of Interest (POI) vector. Locations have a daily maximum and minimum temperature, average wind speed and total precipitation, which impact the traffic and people's movement. In addition, locations consists of several POIs that can be categorized into functions. For instance, one grid cell in $S$ might contain many hotels and few shopping centers, while another grid cell might contain many shopping centers. The distribution of categories of POIs over the space impacts people's movement. Therefore, we define a POI vector $V^l=(v^l_1,v^l_2,...,v^l_n)$, where $v^l_i$ is the number of places in category $i$ at $l$.

\subsection{Problem Statement}

\textbf{Given}: Spatio-temporal field $Z=(S,T)$, historical trip records and weather information in $Z$, POI vectors of $S$, significance threshold $\alpha$, current time $t_c$ and target period $(t_c,t_g]$, \textbf{Find}: All the dispersal events and their (1) Start time $t_e\leq t_g$ of the dispersal event and (2) Demand volume $C^p_{T_g}$, in case of a predicted dispersal event, where $T_g=[t_e,t_g]$. Our objective is to improve accuracy of $t_e$ and $C^p_{T_g}$.

\section{Computational Solution}
\label{solution}
\subsection{Overview}
\begin{figure}
	\centering
	\includegraphics[width=0.45\textwidth]{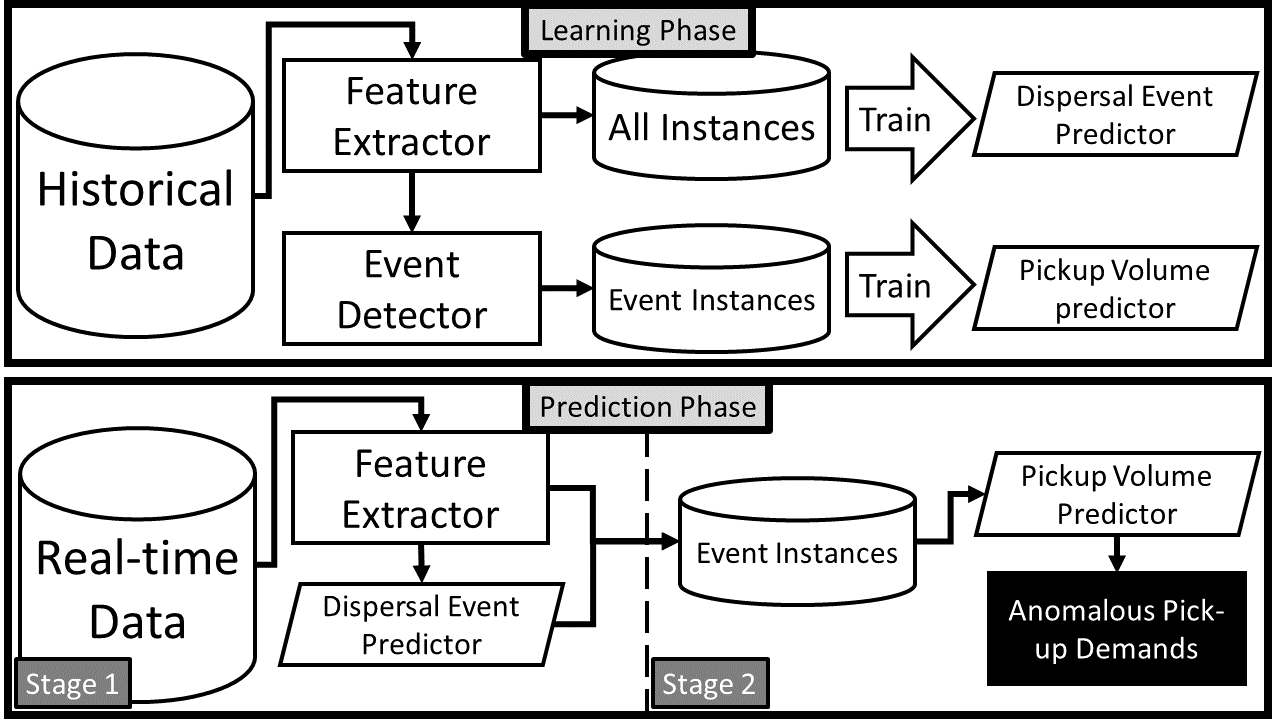}
	\caption{\small Dispersal event prediction framework.}
	\label{diag}
\end{figure}

Per problem statement, we predict (1) start time of dispersal events, (2) demand during dispersal events. We propose the framework in Fig. \ref{diag}. In the learning phase, we extract features from historical data and use them to train an event predictor based on Survival Analysis and a demand predictor. In the prediction phase, we follow two steps. First, we use the event predictor to predict the start time of the event. Then, we predict the pick-up count for the period of the event.

\subsection{Survival Analysis}
Survival analysis analyzes the expected time until an event happens \cite{miller2011survival}. The event could be death or failure, or in this paper, a dispersal event. The analysis is primarily done using the survival function defined as follows:
\begin{equation}
\small
S(t)=Pr(E>t).
\label{surv}
\end{equation}
In Eq. \ref{surv}, $S(t)$ is the probability of the event not happening until $t$ (subject has survived at $t$). Another commonly used function is the hazard function $h(t)$, which is the rate of event at time $t$, given that it has not occurred by then. Hazard function is defined as follows:
\begin{equation}
\small
h(t)=\frac{-S'(t)}{S(t)}.
\end{equation}
$-S'(t)$ is the rate with which $S(.)$ decreases at $t$. It is divided by $S(t)$, the remaining mass of survival probability, because it is conditional to the survival of the subject at $t$. We use this analysis to calculate the remaining time to dispersal events.
\subsection{Feature Extraction}
\label{feature}
To do supervised learning, we need to have a training set with instances of inputs and outputs. In this section, we define the input variables, or the building blocks of the feature vector of the supervised learning framework. Let $(l,t_c)$ be the current location and time. We build the variables through following definitions:
\begin{definition}
	\textbf{Time profile} of $(l,t_c)$ is $Q^l_{t_c}=\langle d_y,d_w,t_c-t_d \rangle$, where $d_y$ and $d_w$ are the day of the year and day of week for $t_c$, and $t_d$ is the first time-step of current day.
\end{definition}

\begin{definition}
	\textbf{Weather profile} of $(l,t_c)$ is $W^l_{t_c}=\langle \omega,\eta,\zeta,\theta_{max},\theta_{min} \rangle$, where $\omega$ is average daily wind speed, $\eta$ is total rain fall of the day, $\zeta$ is total snowfall of the day and $\theta_{max}$ and $\theta_{min}$ are the maximum and minimum temperatures of $l$ at $t_c$.
\end{definition}

\begin{definition}
	\textbf{Daily profile} of $(l,t_c)$ is defined as:
	\begin{equation}
	\small
	M^l_{t_c}=\Bigg\langle\sum_{t\in {[t_d,t_c)}}C^p_{l,t},\sum_{t\in {[t_d,t_c)}}B^p_{l,t},\sum_{t\in {[t_d,t_c)}}C^d_{l,t},\sum_{t\in {[t_d,t_c)}}B^d_{l,t}\Bigg\rangle.
	\end{equation}
\end{definition}
The daily profile is a vector containing the sum of pick-up and drop counts and baselines since the start of current day. It is important, because a gradual gathering during the day can result in an accumulation of people in $l$ at $t_c$, which might not be obvious in individual time-steps. Next, we define the recent profile of $\textbf{x}$.
\begin{definition}
	\textbf{Recent profile} of $(l,t_c)$ is defined as:
	\begin{equation}
	\small
	N^l_{t_c}=\big\langle C^p_{l,t_c-\tau},C^d_{l,t_c-\tau},..., C^p_{l,t_c},C^d_{l,t_c}\big\rangle
	\end{equation}
	where $\tau$ is a parameter.
\end{definition}
The recent profile contains all the pick-up and drop counts of the recent $\tau$ time-steps at current location. 
We define the target profile as follows:
\begin{definition}
	\textbf{Target profile} of $(l,t_c)$ is defined as:
	\begin{equation}
	\small
	G^l_{t_g}=\big\langle B^p_{l,t_c+1},...,B^p_{l,t_g}\big\rangle.
	\end{equation}
	where ${(t_c,t_g]}$ is the target period, i.e. the time period for which we are going to make predictions.
	\label{target}
\end{definition}
The target profile is the expected pick-up counts of the prediction target time period in the future. We define the anomaly profile as follows:
\begin{definition}
	\textbf{Anomaly profile} of $(l,t_c)$ is defined as:
	\begin{equation}
	\small
	F^l_{t_g}=\big\langle LLR(l,t_c+1),...,LLR(l,t_g)\big\rangle.
	\end{equation}
	where ${(t_g-1,t_g]}$ is the target period, same as Definition \ref{target}.
\end{definition}
The anomaly profile is consisted of the anomaly scores of $l$ during the target period based on Eq. \ref{ebp}. These values are available during training, but not during testing. We will use predicted anomaly scores instead, while testing.

\subsection{Building the Training Sets}
As in Fig. \ref{diag}, we train three estimators: survival function estimator ($f_s$), anomaly profile estimator ($f_a$) and dispersal event pick-up predictor ($f_e$). In this section, we describe how their training sets are obtained. We propose to use estimators that are maintain an internal state, such as recurrent neural networks. Thus, the order of instances in the training set matters. This order must match the order of real-time data. Ensuring this requirement is straightforward for $f_a$ and $f_s$, since they are trained using all the instances. However, it is not straightforward for $f_e$, because it is not trained on all instances. Later in this section, we will demonstrate how this requirement is satisfied.

As mentioned earlier, we treat the dispersal event prediction problem as a survival problem. Therefore, the output vector for $f_s$ is the survival probabilities. In this case, the dispersal event is the death event in the survival problem. To this end, the survival function is defined as follows:
\begin{equation}
\small
S(t)=Pr(E^p>t)
\label{survive}
\end{equation}
where $E^p$ is the time of dispersal event. In our proposed framework, we train a model to predict $S(t)$. For location $l$ and time $t_c$, we use the following input vector for $f_s$:
\begin{equation}
\small
\label{xs}
\textbf{x}_s=\big\langle Q^l_{t_c},W^l_{t_c},M^l_{t_c},G^{l}_{t_g},F^{l}_{t_g},V^l,N^i_{t_c}, S(t_c) \big\rangle, i\in l^*.
\end{equation}
We call $l^*$ the surrounding area of $l=(a,b)$ defined as the rectangular area bounded by grid cells $(a-\lambda,b-\lambda)$ and $(a+\lambda,b+\lambda)$, where $\lambda$ is a parameter. Input vector $\textbf{x}_s$ consists of time, weather, daily, target and anomaly profiles and the POI vector of $(l,t_c)$ and the recent profile of $(l^*,t_c)$. $\textbf{x}_s$ plus the current value of the survival function $S(t_c)$.

\begin{figure}
	\centering
	\includegraphics[width=0.22\textwidth]{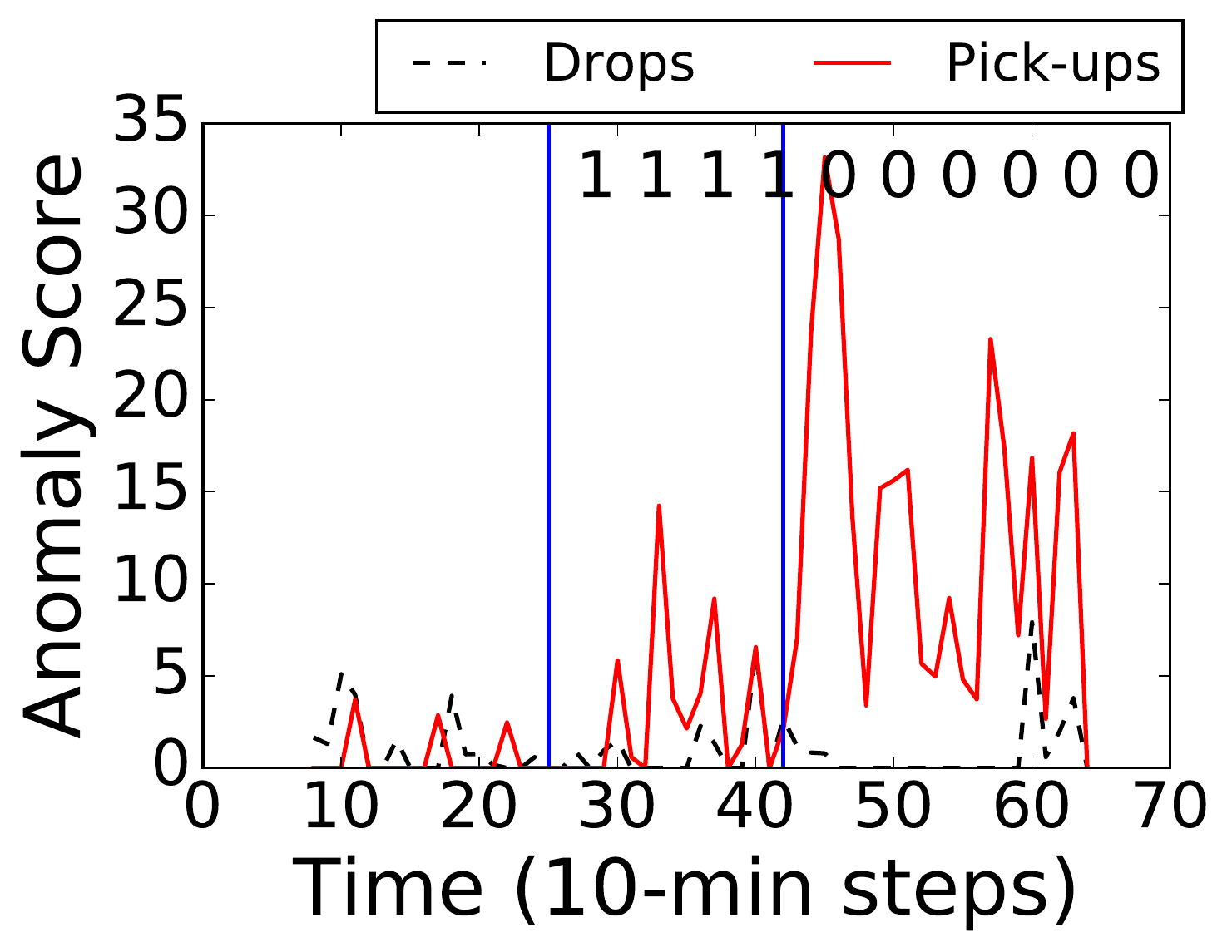}
	\caption{\small An example of the values of the survival function $S(.)$ in case of a dispersal event.}
	\label{example_3}
\end{figure}

For each input vector $\textbf{x}_s$ at location $l$ and time $t_c$, we use the following output vector:
\begin{equation}
\small
\textbf{y}_s=\big\langle S(t_c+1),...,S(t_g) \big\rangle.
\label{ys}
\end{equation}
Ideally, we would like to have a labeled event list for our training phase. However, such lists are not available. Therefore, we use an algorithm to obtain $S(.)$ for a given time and location $(l,t_c)$ by determining if any dispersal event has occurred, or is underway in the future of $t_c$. This procedure is presented in Alg. \ref{pseudo_event}. We put a limit on the length of a dispersal event, assuming the events that are shorter than $e_{min}$ or longer than $e_{max}$ are not interesting. Then, we test every sub-period between $t_c-e_{max}$ and $t_g$ that are longer than $e_{min}$, using Def. \ref{disE}. The survival value will be set to one before and zero after the start of the dispersal event. For example, consider Fig. \ref{example_3}, which shows the dispersal event of Fig. \ref{example_1} (b). The first vertical line is the current time, the second vertical line is the starting time of the dispersal event. The survival function is set to $1$ before the start of the event and is set to zero afterwards.
Alg. \ref{pseudo_event} calculates the survival function. $(t_c-e_{max},t_c+t_g)$ has exponential number of sub-periods. However, we are only interested in the earliest dispersal event, because the survival function will be zero afterwards. Alg. \ref{pseudo_event} takes advantage of this fact and runs in $O(nm)$, where $n$ is the length of time being searched ($end-start$) and $m$ is the number of different lengths sub-periods can have ($e_{max}-e_{min}$).

$\textbf{x}_s$ and $\textbf{y}_s$ are obtained for every spatio-temporal grid cell in $Z$. They constitute the training set for $\textbf{y}_s=f_s(\textbf{x}_s)$ for estimating the survival function. We will discuss how we use $f_s$ to predict the start time of dispersal events.
\begin{algorithm}[h]
	\SetKwInput{Input}{Input}
	\SetKwInput{Output}{Output}
	\small
	\LinesNumbered
	\DontPrintSemicolon
	\BlankLine
	\caption{Calculate survival function (get\_St)}
	\label{pseudo_event}
	\Input{Baselines and counts, current location $l$, current time $t_c$, target time $t_g$}
	\Output{Survival function $S(t)$ where $t\in {(t_c,t_g]}$}
	\BlankLine
	$S(.) \leftarrow \{1\}$; $start \leftarrow t_c - e_{max}$; $end \leftarrow t_g$\;
	\For{$k$ from $e_{max}$ to $e_{min}$}
	{
		\For{$t_0\in [start,end-k]$}
		{
			$t_1 \leftarrow t_0 + k$\;
			\If{$LLR(l,[t_0,t_1])$ is significant and $t_1 > t_c$}
			{
				\For{$t \in [max(t_0,t_c),t_g]$}
				{
					$S(t) \leftarrow 0$\;
				}
				\Return $S(.)$\;
			}
		}
	}
	\Return $S(.)$\;
\end{algorithm}

Although anomaly profile ($F^l_{t_g}$) values are available during training, they are not available during testing, because we do not have the true pick-up counts in the future. While, we train $f_s$ using the true anomaly profile, we have to use predicted anomaly profile in the prediction phase. We use $f_a$ to predict the anomaly profile. The input vector of $f_a$ is denoted as $\textbf{x}_a$ and is shown as follows:
\begin{equation}
\small
\label{xa}
\textbf{x}_a=\big\langle Q^l_{t_c},W^l_{t_c},M^l_{t_c},G^{l}_{t_g},F^{l}_{{(t_c-\tau,t_c]}},V^l,N^i_{t_c} \big\rangle, i\in l^*.
\end{equation}
Where $F^{l}_{{(t_c-\tau,t_c]}}$ is the anomaly profile in the recent time period. Eq. \ref{xa} means that we use the time, weather, daily, recent profiles and the POI vector, in addition to recent anomaly values to predict the future anomaly profile.

Next, we predict the pick-up counts in case of dispersal events, using estimator $f_e$. We use an input vector with the same elements as $\textbf{x}_s$. The output vector of $f_e$ is as follows:
\begin{equation}
\small
\label{ye}
\textbf{y}_e=\big\langle C^p_{l,t_c+1},...,C^p_{l,t_g}\big\rangle
\end{equation}
Although $f_s$ and $f_e$ have the same feature vectors, they are not trained with the same sets. This is a key point in our proposed approach. In the training set of $f_e$ we only include the data instances which correspond to a dispersal event. The reason is we will only use $f_e$ to predict the pick-up counts in case of abnormally high pick-up counts. Thus, we train it with just those instances. Alg. \ref{pseudo_fe} builds the training set for $f_e$. To make sure $f_e$ learns a full cycle of a dispersal event in its internal state, for each event, we include all the instances starting from the time when the event is first observed in the target period (line 5). For example, let $t$ be current time and the target period be $4$ time-steps long. If the survival function is $\langle 1,1,1,0 \rangle$, then the instances of time-steps ${[t,t+4)}$ will be included in the training set (lines 6-9).

\begin{algorithm}
	\SetKwInput{Input}{Input}
	\SetKwInput{Output}{Output}
	\small
	\LinesNumbered
	\DontPrintSemicolon
	\BlankLine
	\caption{Build dataset for $f_e$ (get\_Xe)}
	\label{pseudo_fe}
	\Input{Baselines and counts, time, weather, daily, recent, target and anomaly profiles, POI vectors, $Z=(S,T)$}
	\Output{Training set $\textbf{X}_e$ and $\textbf{Y}_e$ for $f_e$}
	\BlankLine
	$\textbf{Y}_e$ = An empty list; $\textbf{X}_e$ = An empty list\;
	\For{$l \in S$}
	{
		\For{$t\in T$}
		{
			$S(t)$ = get\_St($l$,$t$)\;
			\If{$S(t_g) = 0$ AND $S(t_g-1) = 1$}
			{
				\For{$t_c \in [t,t_g]$}
				{
					$\textbf{x}_e$ = $(Q^l_{t_c},W^l_{t_c},M^l_{t_c},G^{l}_{t_g},F^{l}_{t_g},V^l,N^i_{t_c}), i\in l^*$\;
					$\textbf{y}_e$ = $(C^p_{l,t_c+1},...,C^p_{l,t_g})$\;
					$\textbf{X}$.push\_back($\textbf{x}_e$); $\textbf{Y}$.push\_back($\textbf{y}_e$)\;
				}
				$t=t_g$\;
			}
		}
	}
	\Return $\textbf{X}_e$, $\textbf{Y}_e$\;
\end{algorithm}
\subsection{DILSA: Dispersal Event Prediction using Survival Analysis}
\begin{figure}
	\centering
	\includegraphics[width=0.35\textwidth]{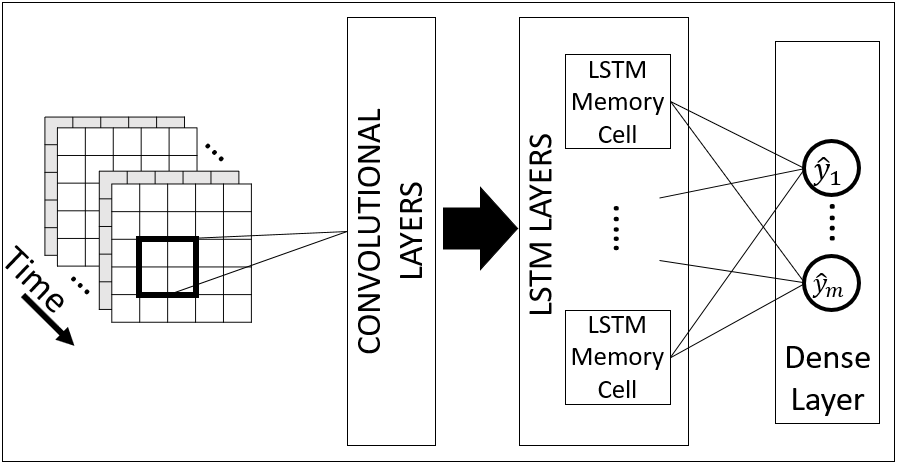}
	\caption{\small Deep Learning structure used to learn spatial and temporal dependencies.}
	\label{ann}
\end{figure}

The training sets built in the previous section contain temporal and spatial dependencies. Thus, we use a Deep Artificial Neural Network that uses Convolutional layers to capture spatial dependencies and LSTM layers to capture temporal dependencies. Fig. \ref{ann} shows the employed structure.

First step in our framework is to obtain the anomaly profile of the target period using $f_a$, to be used in the input vector of $f_s$. Then, $1-S(.)$ is the estimated cumulative probabilities of event. Assuming $S(0)=1$, we calculate the probability of event at future time using the hazard function:
\begin{equation}
\small
H(t)=\frac{S(t-1)-S(t)}{S(t)}.
\label{haz}
\end{equation}
Eq. \ref{haz} calculates the cumulative hazard of event happening between $t-1$ and $t$ given that it has not happened as of $t-1$. This value is calculated by dividing the amount of drop in the survival function from time $t-1$ to $t$, by the total remaining amount, which is $S(t)$, given that $S(.)$ monotonically decreases. We predict an event, when value of $H(.)$ exceeds a threshold $\gamma$, which is tuned using a tuning set.

Once a dispersal event is predicted, we predict the pick-up count for the event using $f_e$. Since our estimators maintain an internal state, we must make predictions in the same order as training. This is not a problem for estimators $f_a$ and $f_s$, because they were trained using all the instances, which is the same order of real-time data. To train $f_e$, Alg. \ref{pseudo_fe} establishes a specific order that must also be followed in the prediction phase. In the training phase, we included instances when the start of the event first appears in the target period, i.e. the survival function turns to $0$ in the last time-step of the target period ($S(t_g)=0$ and $S(t_g-1)=1$). Therefore, we must start predicting the pick-ups using $f_e$ once Eq. \ref{haz} predicts the last time-step of the target period to be $0$. However, Eq. \ref{haz} might not predict the occurrence of the event until the start time gets closer. In such a case, $f_e$ will not have correct internal state. Therefore, to bring $f_e$ to its correct internal state, we feed the input vectors of previous time-steps to $f_e$ before the input vector of current time. For example, suppose we are at time $t_c$ and target time period is $4$ time-steps long. Then we predict a dispersal event at time $t_c+2$. For $f_e$ to make predictions for $t_c+2$ and $t_c+3$, we feed the input vectors of time $t_c-2$, then $t_c-1$ to $f_e$. Now $f_e$ has the correct internal state to make predictions.

Alg. \ref{pseudo_predictor} shows the proposed dispersal event demand predictor. First, the anomaly profile is obtained and used to predict the survival function (lines 1-2). Then $H(.)$ is calculated for future periods and compared with threshold $\gamma$ to predict the dispersal events (lines 4-9). A value of $1$ in $\hat{\textbf{y}}_s[t]=1$ means a dispersal event is predicted for $t$ time-steps after current time. In case of a predicted event, the internal state of $f_e$ is corrected and pick-up counts are predicted (line 10-13).
\begin{algorithm}
	\SetKwInput{Input}{Input}
	\SetKwInput{Output}{Output}
	\small
	\LinesNumbered
	\DontPrintSemicolon
	\BlankLine
	\caption{\small Dispersal event predictor (DILSA)}
	\label{pseudo_predictor}
	\Input{Estimators $f_s(.)$, $f_a$ and $f_e(.)$, current time $t_c$, target time $t_g$, threshold $\gamma$}
	\Output{Predicted dispersal events $\hat{\textbf{y}}_s$, predicted counts of the predicted events $\hat{\textbf{y}}_e$}
	\BlankLine
	\For{$l\in S$}
	{
		$\hat{\textbf{y}}_s[l]=\{0\}$; $\hat{\textbf{y}}_e[l]=\{-1\}$; $\textbf{x}_a$=construct\_$\textbf{x}_a$($l$,$t_c$)\;
		$\hat{\textbf{F}}$=$f_a(\textbf{X}_a)$; $\textbf{x}_s$=construct\_$\textbf{x}_s$($l$,$t_c$,$\hat{\textbf{F}}$); $St=f_s(\textbf{x}_s)$\;
		is\_event=\textbf{False}\;
		\For{$i\in{[1,t_g-t_c)}$}
		{
			$H=(S(i-1)-S(i))$/$S(i)$\;
			\If{$H \geq \gamma$}
			{
				\For{$j\in{[i,t_g)}$}
				{
					$\hat{\textbf{y}}^s_j[l]=1$\;
				}
				is\_event=\textbf{True}; event\_time=i; break\;
			}
		}
		\If{is\_event}
		{
			Correct the internal state of $f_e$\;
			$\textbf{x}_e$=construct\_$\textbf{x}_e$($l$,$t_c$,$\hat{\textbf{F}}$)\;
			$\hat{\textbf{y}}_e[l]=f_e(\textbf{x}_e)$\;
		}
	}
	Return $(\hat{\textbf{y}}^s,\hat{\textbf{y}}^e)$\;
\end{algorithm}

\section{Evaluations}
\label{evaluation}
\subsection{Settings and Baseline Solutions}
\label{settings}
We use the trip records of Yellow Taxis in New York City from years 2014, 2015 and 2016. This dataset contains the pick-up and drop locations and is released by New York City Mayor's Office \footnote{\tiny\url{https://opendata.cityofnewyork.us/overview/}}. The weather data is obtained from the National Centers for Environmental Information \footnote{\tiny\url{https://www.ncei.noaa.gov/}} from two weather stations, Central Park and the La Guardia Airport. The Point of Interest data is obtained from Google Maps Places API \footnote{\tiny\url{https://developers.google.com/places/}}, which assigns POIs into one or more of 129 categories. We partition the New York City area into a grid of $32\times32$ with cell size of $400\times400$ meters. We use 30-minute time-steps. Every record is mapped into the grid to obtain counts and baselines. The values of weather profile for each spatio-temporal grid cell is an average of the measurements reported by the two stations, weighted inversely by their distance. We train the models using year 2014 and evaluate on 2015 and 2016. All datasets are standardized by subtracting the minimum and dividing by the maximum value of each feature. The test sets are standardized using parameters from the training set. Table \ref{params} shows our parameter settings.

\begin{table}
	\small
	\centering
	\caption{\small Parameter settings.}
	\label{params}
	\begin{tabular}{llllll}
		\toprule
		$\lambda$ & $\alpha$ & $t_g-t_c$ & $\tau$ & $e_{min}$ & $e_{max}$\\
		\midrule
		4 & 0.001 & 5 hrs & 5 hrs & 30 min & 5 hrs \\
		\bottomrule
	\end{tabular}
\end{table}

In table \ref{params}, $t_g-t_c$ is the duration of the target period. Our Deep Learning Network uses 4 convolutional layers with window size of $9\times 9$, $2$ LSTM layers of $69$ memory cells and 10 output nodes. We compare $f_e$ with state-of-the-art deep learning method for taxi pick-up prediction, \textbf{DMVST-Net} \cite{yao2018deep}. Moreover, we use three additional baselines for comparison. First baseline is simple thresholding of the survival function instead of Eq. \ref{haz}, i.e. if the survival function drops below a threshold ($\sigma$), the event is predicted. We call this baseline \textbf{DIL}. We tune both $\gamma$ and $\sigma$, using a week's data in 2015 ($\gamma=2.95$ and $\sigma=0.1$). We also compare with Multi-Layer Perceptron (\textbf{MLP}) and Logistic Regression (\textbf{LgR}) models.

The estimators were trained using the stochastic gradient descent method proposed by \cite{kingma2014adam}, with $20$ epochs for $f_a$ and $f_s$ and $40$ epochs for $f_e$.

\subsection{Case Studies}
\label{case}
We apply the proposed method to a full dataset from 2016. Here, we present two of the predicted events. 

\begin{figure}[t]
	\centering
	\subfigure[Event location.]{\includegraphics[width=0.12\textwidth]{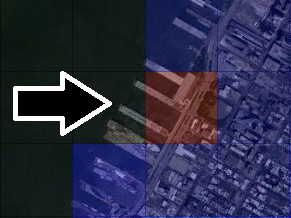}}
	\subfigure[Case 1 survival curve.]{\includegraphics[width=0.15\textwidth]{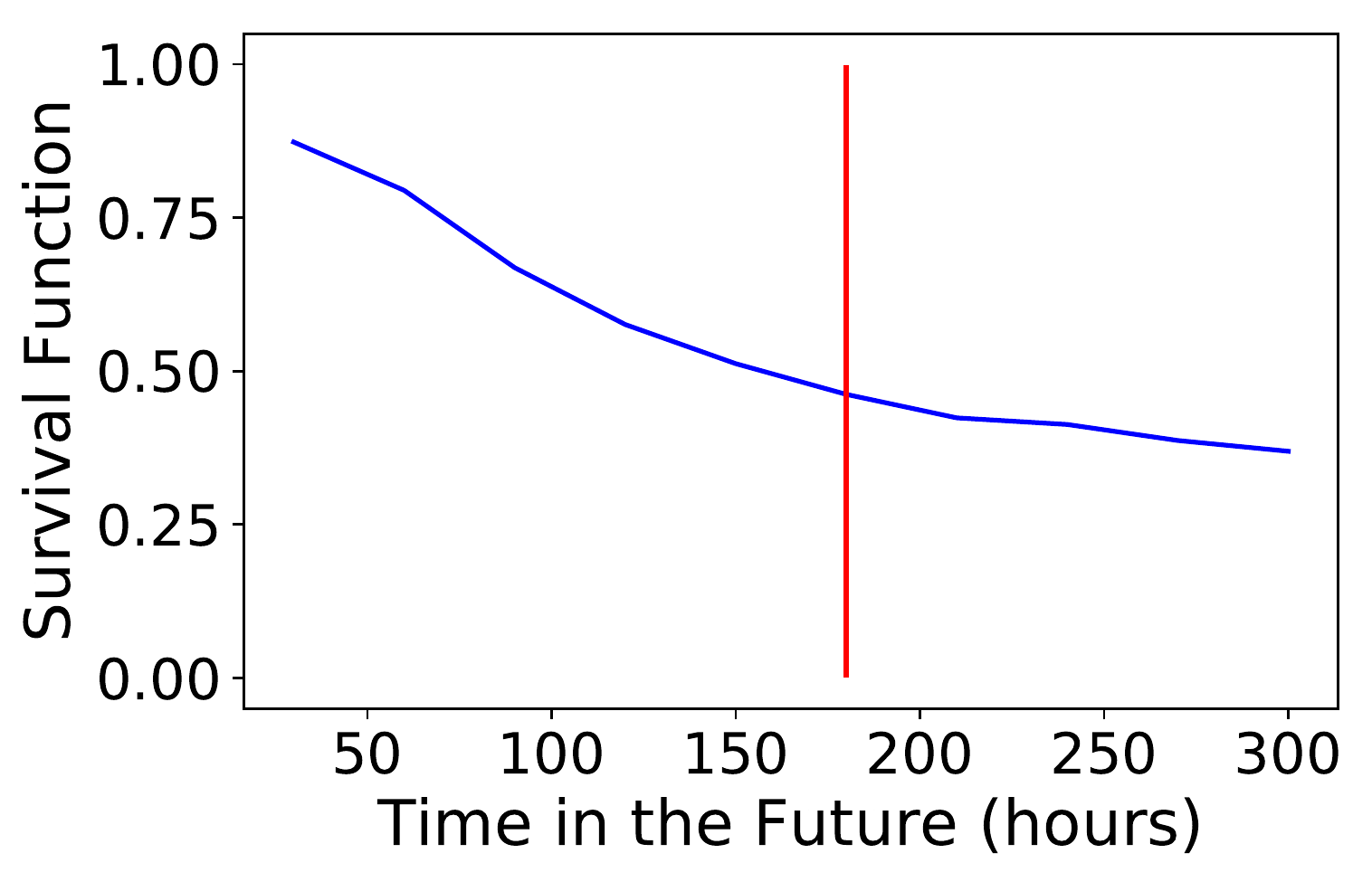}}
	\subfigure[Predicted pick-up counts vs. true counts.]{\includegraphics[width=0.19\textwidth]{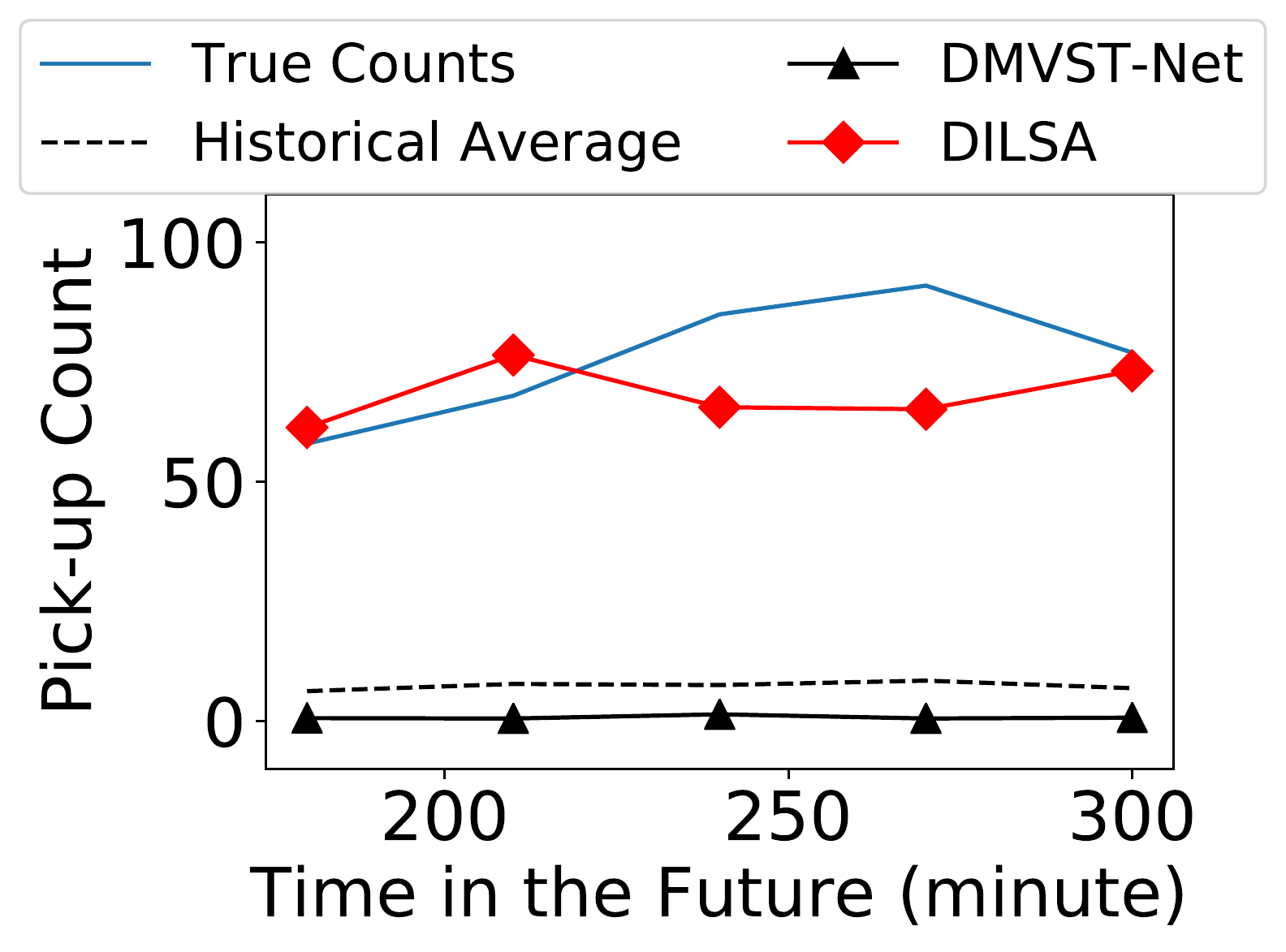}}
	\caption{\small First case study (best viewed in color).}
	\label{fig_case1}
\end{figure}

On March $19^{th}$, 2016, we predicted a dispersal event at 1:00 PM around an exhibition center in Pier 92/94 in Manhattan. We predict the event $2.5$ hours before (at 11:30 AM). Public records show a home design exhibition at the time \footnote{\tiny\url{https://architecturaldigest.com/story/architectural-digest-design-show-video}}. Fig. \ref{fig_case1} (b) shows the predicted survival curve at 11:30 AM. The red vertical line is the predicted time of the dispersal event, which is inferred by Alg. \ref{pseudo_predictor}. Fig. \ref{fig_case1} (c) shows the predicted counts by the baseline and the proposed method. The proposed method successfully predicts the increase, while DMVST-Net stays close to the historical average.

\begin{figure}[t]
	\centering
	\subfigure[Event location.]{\includegraphics[width=0.12\textwidth]{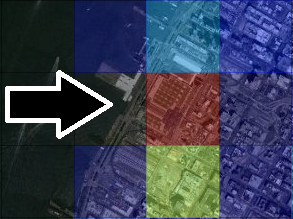}}
	\subfigure[Case 2 survival curve.]{\includegraphics[width=0.15\textwidth]{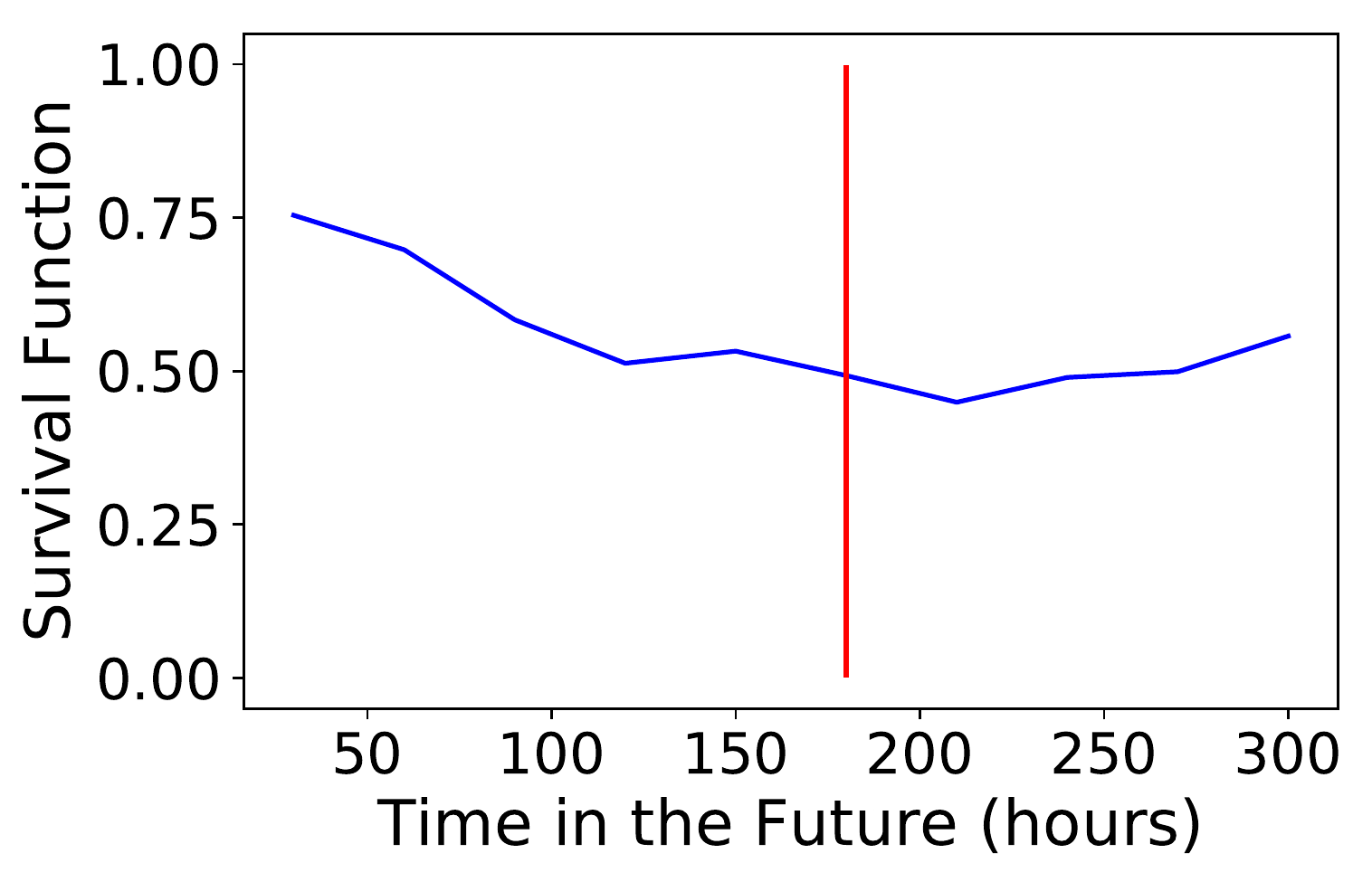}}
	\subfigure[Predicted pick-up counts vs. true counts.]{\includegraphics[width=0.19\textwidth]{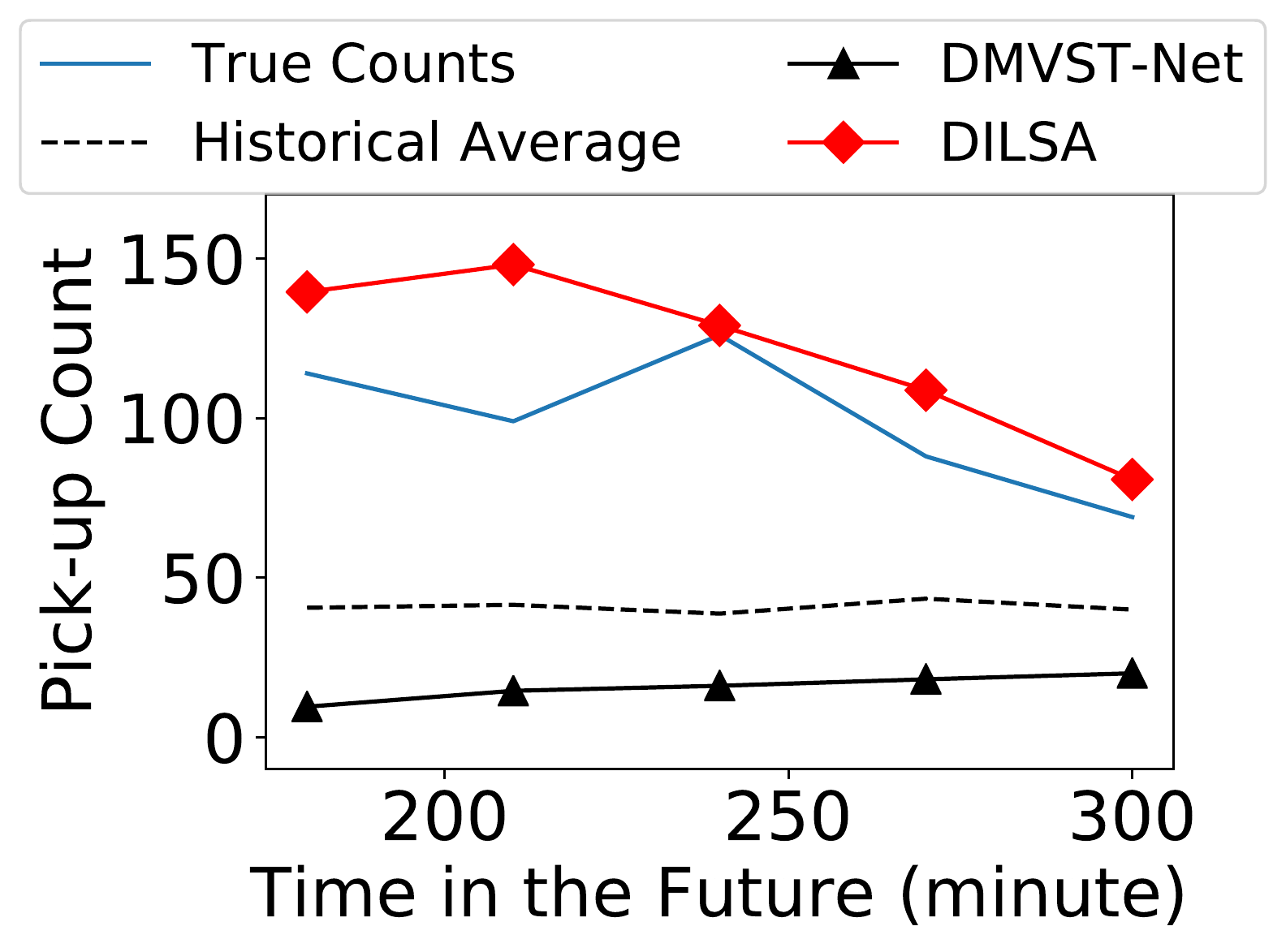}}
	\caption{\small Second case study (best viewed in color).}
	\label{fig_case2}
\end{figure}

We also predicted a dispersal event around 12:30 PM on June $26^{th}$, 2016, at Jacob K. Javis Convention Center, $2.5$ hours before. Public records show there was a food show at the convention center \footnote{\tiny\url{https://specialtyfood.com/news/article/2016-summer-fancy-food-show-largest-ever/}}. Fig. \ref{fig_case2} (b) shows the predicted survival curve and the event prediction time, indicated by the vertical red line. Fig. \ref{fig_case2} (c) shows the proposed method out-performs DMVST-Net in predicting the pick-up counts in this case. Fig. \ref{fig_case1} (a) and \ref{fig_case2} (a) show heatmaps of LLR scores based on the true counts in the predicted periods. The black arrows show the verified locations. The figures show a clear hotspot of pick-ups. Overall, these two case studies demonstrate examples of DILSA successfully predicting dispersal events and their corresponding demand.

\subsection{Experiments}
In this section, we first evaluate the prediction performance of DILSA, i.e. the performance of Alg. \ref{pseudo_predictor} to predict events. We compare our results with four baselines. Baseline DMVST-Net predicts taxi demand. We apply Def. \ref{disE} to the predicted value to determine if there is a dispersal event. Table \ref{methods} shows that DILSA out-performs all the baselines in terms of F1-score ($0.7$) and time error ($18$ minutes). Time error is the average difference between the true start time and predicted start time of the correctly predicted events. A prediction is considered a true positive if the predicted event period overlaps with the true event period. The results show the proposed survival analysis method predicts dispersal events with high accuracy. Although DMVST-Net demonstrates high precision, its recall is extremely low, meaning the regular patterns fail to predict accurately in case of abnormally high demand. Moreover, the results show using the cumulative hazard function of Eq. \ref{haz} in Alg. \ref{pseudo_predictor} has a considerable impact on model's performance.

\begin{table}[h]
	\centering
	\small
	\caption{\small Performance comparison, DILSA vs. baselines.}
	\label{methods}
	\begin{tabular}{clllll}
		\toprule
		& \textbf{DILSA} & DIL & DMVST-Net & MLP & LgR\\
		\midrule
		Precision & 0.6 & 0.6 & 0.9 & 0.5 & 0.3\\
		Recall & \textbf{0.9} & 0.8 & 0.04 & 0.6 & 0.3\\
		F1-score & \textbf{0.7} & 0.7 & 0.08 & 0.5 & 0.3\\
		\hline
		Time error\\(min.) & \textbf{18.6} & 29.1 & 60 & 59.6 & 80.9\\
		\bottomrule
	\end{tabular}
\end{table}

\begin{figure}
	\centering
	\subfigure[MAE.]{\includegraphics[width=0.2\textwidth]{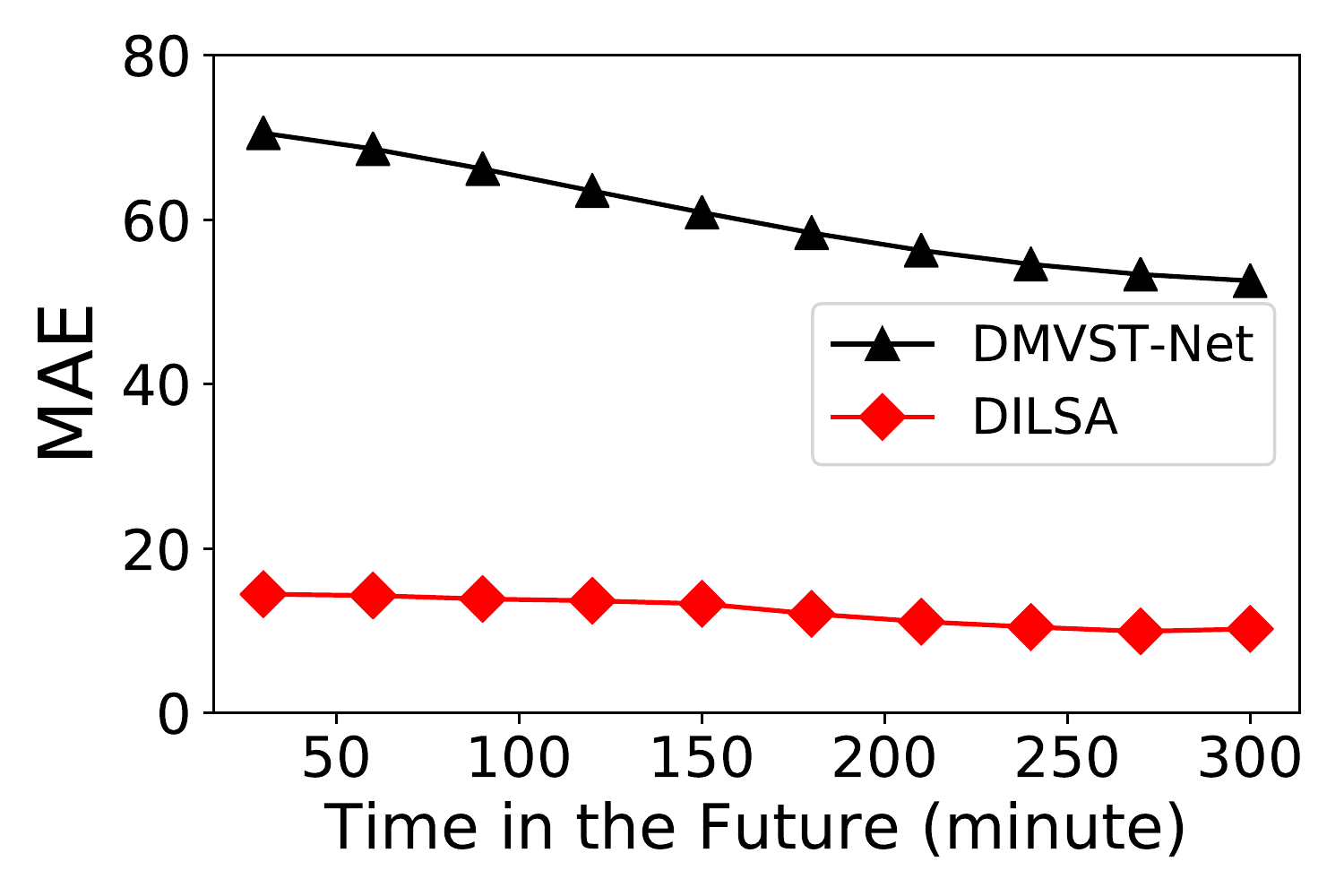}}
	\subfigure[MAPE.]{\includegraphics[width=0.2\textwidth]{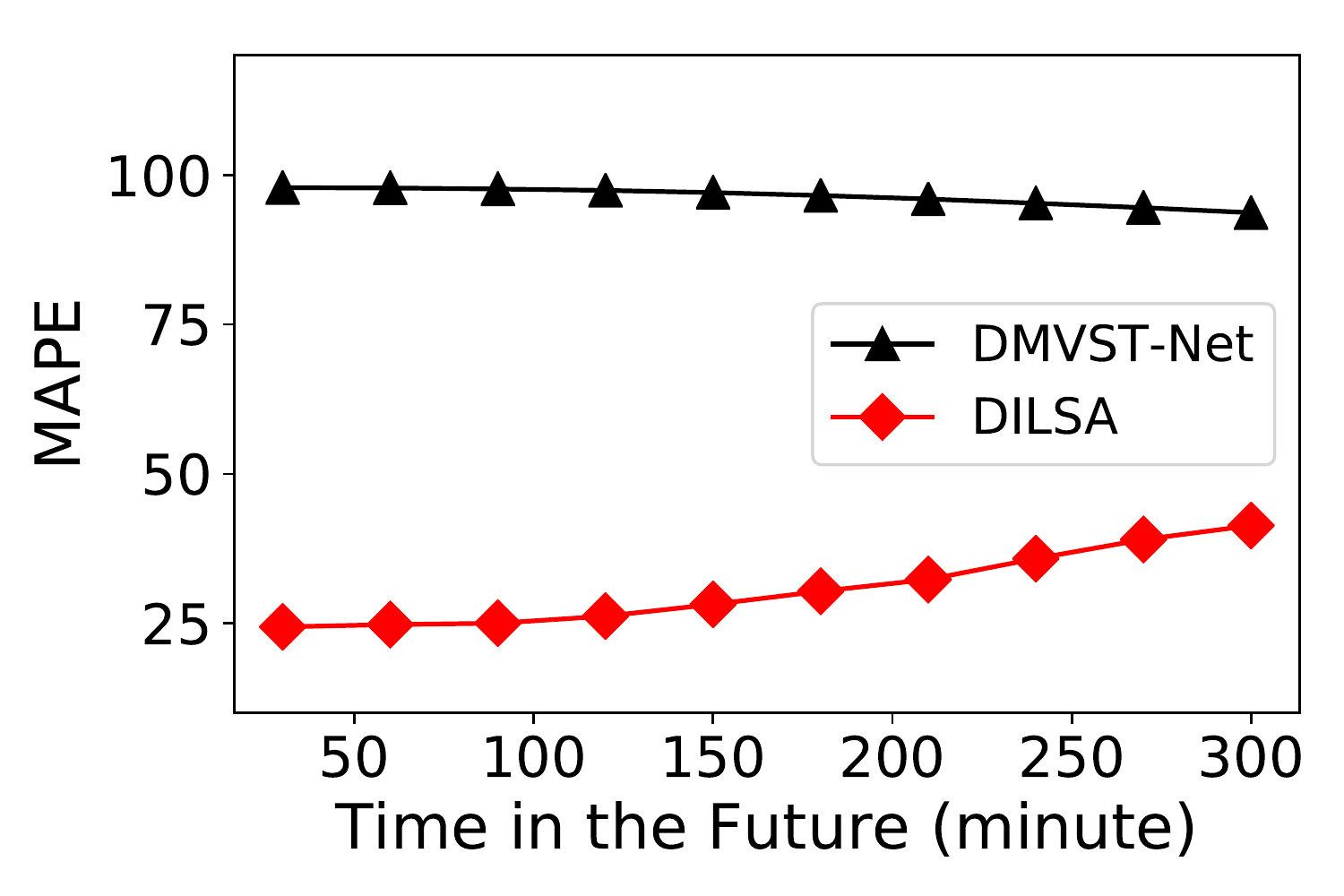}}
	\caption{\small Performance of the proposed pick-up counts predictor vs. baselines, on events.}
	\label{exp3}
\end{figure}

Second, we compare the demand predictor $f_e$ to DMVST-Net in case of dispersal events. The baseline was trained on the same period as the previous experiment. $f_e$ was trained on the dataset obtained using Alg. \ref{pseudo_fe} on the same period of time in 2014. Fig. \ref{exp3} shows Mean Absolute Error (MAE) and Mean Absolute Percentage Error (MAPE) in future time-steps. Fig. \ref{exp3} shows our proposed method out-performs the baseline in case of a dispersal event. This experiment shows methods proposed to capture the regular pattern of taxi demand are not reliable in case of dispersal events.

Lastly, we evaluate the impact of different features on the performance of the models. We use Root Mean Squared Error (RMSE) as the measure. The x-axis represents future time-steps. The letters R, D and P represent the Recent and Daily profiles and the POI vector. The results show including the POI vector reduces the error. Including the daily profile does not have a significant effect on $f_e$ while improves the performance of survival function predictor $f_s$.

\begin{figure}
	\centering
	\subfigure[Survival function predictor.]{\includegraphics[width=0.2\textwidth]{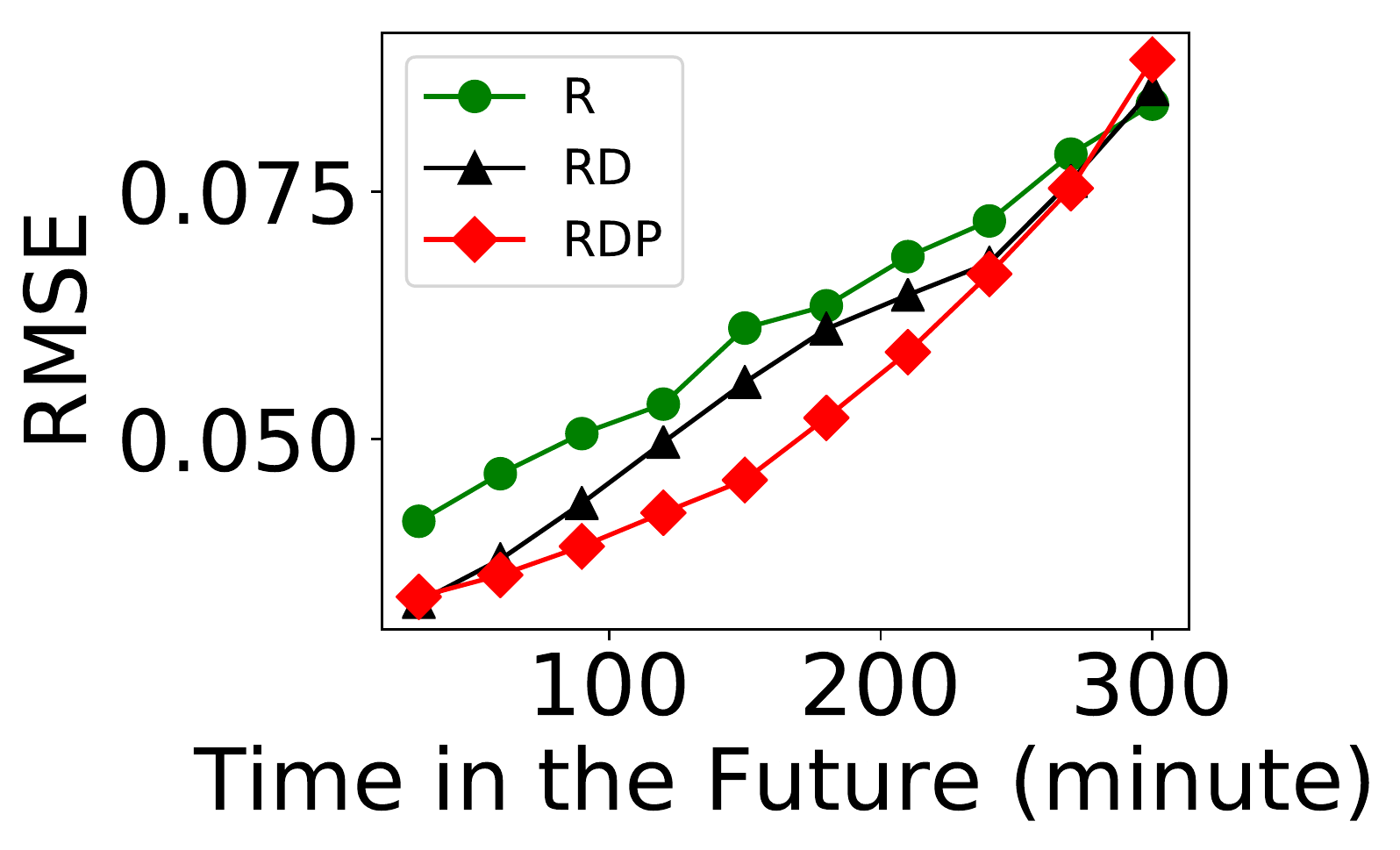}}
	\subfigure[Abnormal demand predictor.]{\includegraphics[width=0.2\textwidth]{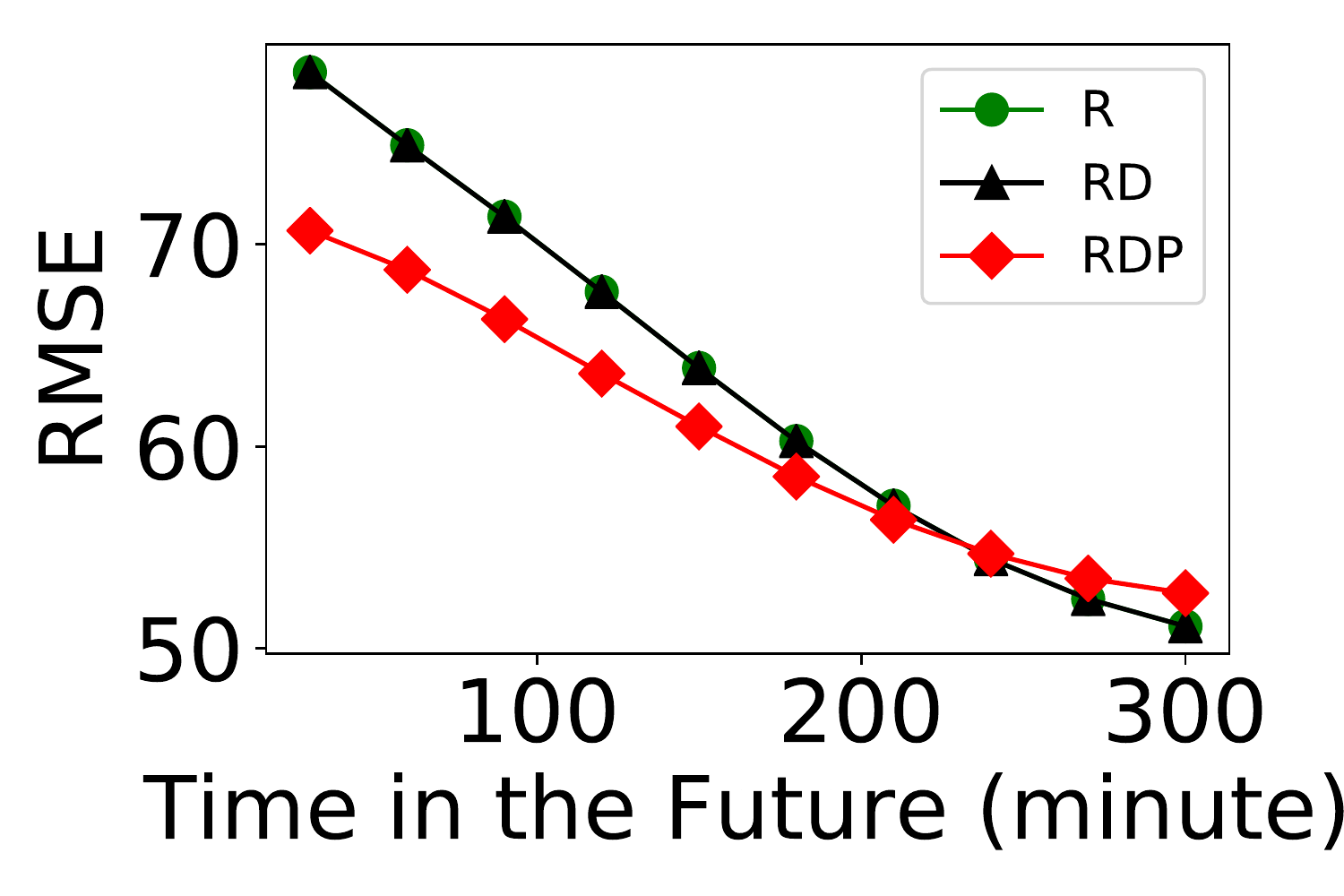}}
	\caption{\small Impact of choice of features on accuracy.}
	\label{exp4}
\end{figure}

\section{Conclusions}
\label{conclusion}
In this paper we solved the problem of predicting dispersal events where a large number of people leave the same area in a short period. Predicting such events has managerial and business value for various stakeholders. We solved the problem as an abnormally high demand prediction problem. The taxi demands in unexpected dispersal events deviate from regular patterns and violate assumptions made by previous techniques (e.g., auto-correlation, periodic). In this paper we argued that dispersal events follow a complex pattern of trips and other related features. We formulated and learned such patterns to predict dispersal events. We formulated the dispersal event prediction as a survival analysis problem and proposed a two-stage framework (DILSA), where a supervised model predicted the probability of ``death'', i.e., the dispersal event. The demand was then predicted in case of a predicted event. We conducted extensive case studies and experiments on a real dataset from 2014-2016. Our method out-performed the baselines and predicted dispersal events with F1-score of $0.7$ and time error of $18$ minutes.

\section{Acknowledgments}
This work is partially supported by the NSF under Grant Number IIS-1566386. We gratefully acknowledge the support of NVIDIA Corporation with the donation of the Titan Xp GPU used for this research. Yanhua Li is partly supported by NSF grant CNS-1657350, CMMI-1831140, and an industrial grant from DiDiChuxing Research.

\bibliography{references}
\bibliographystyle{aaai}

\end{document}